\documentclass[lettersize,journal]{IEEEtran}
\usepackage{amsmath,amsfonts}
\usepackage{algorithmic}
\usepackage{algorithm}
\usepackage{array}
\usepackage{amsmath}
\usepackage{amssymb}
\usepackage{booktabs}
\usepackage[caption=false,font=normalsize,labelfont=sf,textfont=sf]{subfig}
\usepackage{epsfig}
\usepackage{multirow}
\usepackage{multicol}
\usepackage{textcomp}
\usepackage{stfloats}
\usepackage{times}
\usepackage{url}
\usepackage{verbatim}
\usepackage{graphicx}
\usepackage{cite}
\usepackage[colorlinks,
            linkcolor=blue,
            anchorcolor=blue,
            citecolor=blue]{hyperref}

\usepackage{multirow}
\usepackage[table,xcdraw]{xcolor}
\begin{document}

\title{Disentangled Generation Network for Enlarged License Plate Recognition  and A Unified Dataset}

\author{Chenglong Li, Xiaobin Yang, Guohao Wang, Aihua Zheng, Chang Tan, Ruoran Jia, and Jin Tang
\thanks{This work is partly supported by National Natural Science Foundation of China (No. 62076003), and Anhui Provincial Key Research and Development Program (No. 202104d07020008).

C. Li and A. Zheng are with Information Materials and Intelligent Sensing Laboratory of Anhui Province, Anhui Provincial Key Laboratory of Multimodal Cognitive Computation, School of Artificial Intelligence, Anhui University, Hefei 230601, China. (Email: lcl1314@foxmail.com, ahzheng214@foxmail.com).

X. Yang, G. Wang and J. Tang are with Information Materials and Intelligent Sensing Laboratory of Anhui Province, Key Lab of Intelligent Computing and Signal Processing of Ministry of Education, Anhui Provincial Key Laboratory of Multimodal Cognitive Computation, School of Computer Science and Technology, Anhui University, Hefei 230601, China. (Email: yxiaobin2016@163.com, auroraveil@qq.com, tangjin@ahu.edu.cn)

C. Tan and R. Jia are with iFLYTEK Co., Ltd., Hefei 230088, China.  (Email: changtan2@iflytek.com, rrjia@iflytek.com ).
}
}

\markboth{Journal of \LaTeX\ Class Files,~Vol.~14, No.~8, August~2021}%
{Shell \MakeLowercase{\textit{et al.}}: A Sample Article Using IEEEtran.cls for IEEE Journals}


\maketitle

\begin{abstract}
License plate recognition plays a critical role in many practical applications, but license plates of large vehicles are difficult to be recognized due to the factors of low resolution, contamination, low illumination, and occlusion, to name a few. 
To overcome the above factors, the transportation management department generally introduces the enlarged license plate behind the rear of a vehicle. 
However, enlarged license plates have high diversity as they are non-standard in position, size, and style. Furthermore, the background regions contain a variety of noisy information which greatly disturbs the recognition of license plate characters.
Existing works have not studied this challenging problem.
In this work, we first address the enlarged license plate recognition problem and contribute a dataset containing 9342 images, which cover most of the challenges of real scenes. 
However, the created data are still insufficient to train deep methods of enlarged license plate recognition, and building large-scale training data is very time-consuming and high labor cost. 
To handle this problem, we propose a novel task-level disentanglement generation framework based on the Disentangled Generation Network (DGNet),
which disentangles the generation into the text generation and background generation in an end-to-end manner to effectively ensure diversity and integrity,
for robust enlarged license plate recognition. Extensive experiments on the created dataset are conducted, and we demonstrate the effectiveness of the proposed approach in three representative text recognition frameworks.
\end{abstract}


\section{Introduction}
\begin{figure}[t]
\begin{center}
\includegraphics[width=\linewidth]{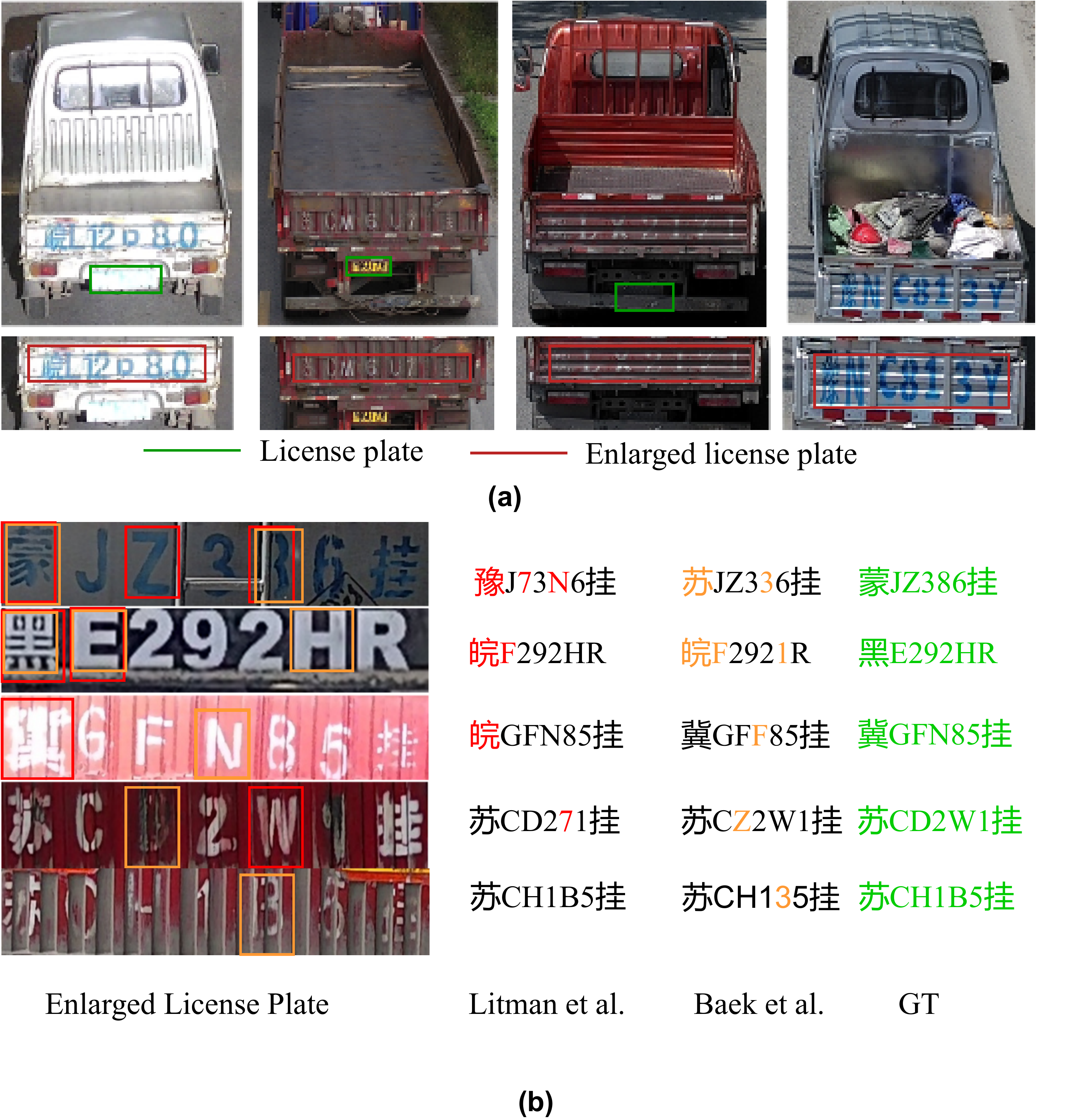}
\end{center}
\caption{
(a) Comparison of enlarged license plates with standard license plates. Enlarged license plate is more suitable than standard license plate in identifying large vehicles under challenging scenarios. (b) Recognition results of the enlarged license plates by the state-of-the-art recognition methods, including Litman et al.\cite{litman2020scatter} and Baek et al.\cite{baek2019wrong}.
%
The results show that the task of enlarged license plate recognition is a challenging task.
}
\label{fig:challenge}
\end{figure}

\IEEEPARstart{L}{icense} plate recognition is an important problem in the field of computer vision and plays a critical role in many practical applications, such as traffic safety, vehicle management, and urban security. 
However, as shown in Fig.~\ref{fig:challenge} (a), when encountering low resolution, contamination, low illumination, and occlusion, the license plates of large vehicles are difficult to be recognized. In such scenarios, it is difficult to manage large vehicles.

According to the requirements of the traffic management department, the rear of large vehicles needs to be painted with enlarged license plates to handle shortcomings of the standard license plates.
In many practical scenarios, the enlarged license plates are easier to be captured by surveillance cameras, and thus the enlarged license plate recognition plays a significant role in identifying large vehicles.
However, enlarged license plate recognition is a very challenging problem. As shown in Fig.~\ref{fig:challenge} (b), on one hand, due to the lack of standard painting requirements in position, size, and style, enlarged license plates are highly diverse. On the other hand, the background contains a variety of noisy information, which has a serious impact on enlarged license plate recognition. Unfortunately, no existing work studies this challenging problem.

In this work, we launch this new task of enlarged license plate recognition, aiming to answer the following two questions: 1) How to create a unified benchmark dataset to promote the research and development of enlarged license plate recognition? 2) How to improve the performance of deep recognizers when training data are insufficient?

{\flushleft \bf Benchmark dataset}.
To establish a unified benchmark dataset, we collect 9342 images, which contain enlarged license plates in 18 provinces in China. This dataset covers most of the real challenges of enlarged license plate recognition such as low resolution, contamination, and occlusion, as shown in Fig.~\ref{fig:challenge} (a). Due to the special properties of enlarged license plates, it is hard to collect balanced data for all provinces. Therefore, the issue of long-tail distribution is a key challenging factor in enlarged license plate recognition. 
To facilitate the training and evaluation of different algorithms, we split the dataset into training and testing sets.
In specific, the testing set occupies about 20~\% of the whole dataset and the remaining data are used as the training set.
We adopt the enlarged license plate recognition accuracy and the character recognition accuracy as our evaluation indicators for different methods.

The huge and diverse challenges such as low illumination and occlusion seriously affect the performance of enlarged license plate recognition.
To facilitate the challenge-based performance analysis, we annotate 10 challenges for each image, including inclined angle, abnormal illumination, different spacing, size variation, blur, abrasion, background clutter, 
non-standard character, double-row plate, and occlusion. 

{\flushleft \bf Disentangled generation network}.
Enlarged license plates have high diversity as they are non-standard in position, size, and style, and background regions contain a variety of noisy information.
Therefore, the created dataset is hard to cover all real challenges and thus insufficient to train deep recognition networks. Moreover, building a large-scale training dataset is very time-consuming and high labor cost. 
To handle this problem, we propose to synthesize large-scale training data to simulate real scenarios.

Many researchers~\cite{cheng2018aon,gupta2016synthetic,jaderberg2014synthetic,wang2017adversarial} propose to generate synthetic images in natural scenes to improve recognition performance. 
For example, 
Wang et al.~\cite{wang2017adversarial} introduce W-Distance~\cite{arjovsky2017towards} in the training process of CycleGAN, and can synthesize a large number of standard license plates, improving the recognition performance. 
However, it is easy to mode collapse, and thus difficult to generate enlarged license plates with various styles. 
Luo et al.~\cite{luo2021separating} introduce an additional recognizer to supervise the generator to ensure the integrity of generated characters. 
However, as discussed above, both enlarged license plates and background present high diversity, which greatly degrades the performance of the recognizer, since the integrity of generated enlarged license plates might not be ensured.

To handle these problems, we propose a novel task-level disentanglement generation framework based on the Disentangled Generation Network (DGNet),
which disentangles the generation into the text and background generations respectively to effectively ensure the diversity and integrity, for robust enlarged license plate recognition.

The effectiveness of disentangle generation is validated in other tasks~\cite{DRIT_plus, huang2018multimodal,yi2020bsd}, but there is no relevant work on license plate recognition. Inspired by the disentangle generation, we propose to disentangle the generation into two processes, including text generation and background generation, to address the problem of mode collapse. 
In specific, 
existing disentangle generation networks~\cite{DRIT_plus,huang2018multimodal} usually extract content and background from source the domain and the target domain respectively, and then combine the content and background to generate target image. 
However, enlarged license plates have high diversity, which makes it difficult to accurately separate the text and background from an
image.
Therefore, we propose a task-level disentanglement generation framework that disentangles the task of enlarged license plate generation into two processes, including text generation and background generation, in an end-to-end manner.

For text generation, we first collect a series of license plate character images. Then, we combine these images into a unified blue background image. To increase the diversity of text images, we augment these character images by changing the attributes of size, shape, and position, etc. More importantly, the changes of these attributes are completely retained in synthesized enlarged license plates, and the diversity is thus enhanced.

For background generation, the complex and diverse backgrounds of enlarged license plates are important factors that affect the performance of recognition methods. To better simulate the real data, we construct a complex and diverse background template set, which contains almost all the backgrounds of enlarged license plates in real scenes. 
Based on this set, we can randomly combine background templates and text images to generate high-quality synthesized enlarged license plates.

Instead of using recognition methods to supervise generation in existing works~\cite{luo2021separating}, we use a mask image to supervise the generation by introducing a mask constraint loss, which is calculated by the average absolute error between the mask image and the output image. Herein, the mask image is obtained by subtracting the blue background image from the text image. The designed loss helps us to effectively ensure the integrity of generated enlarged license plates.

{\flushleft \bf Contributions}.
The main contributions of this work can be summarized as follows.
\begin{itemize}
\setlength\itemsep{-.3em}
\item We propose a new task called enlarged license plate recognition, which is challenging but with high value in real-world applications. To promote the research and development of this research field, we contribute a dataset containing 9342 images, which cover
most of the challenges of real scenes, for enlarged license plate recognition. 

\item To provide a large-scale training data while avoiding time consuming and high labor cost, we propose a novel task-level disentanglement generation framework, which disentangles the enlarged license plate generation into the
text generation and background generation in an end-to-end manner. 

\item We design a series of strategies to ensure the diversity and integrity of generated enlarged license plates. On one hand, we combine a set of augmented text images and a constructed background templates to enhance the diversity. On the other hand, we design a  mask constraint loss based on the mask images to ensure the integrity.

\item We evaluate the generated enlarged license plates in three representative text recognition methods on the created dataset, and the results demonstrate the effectiveness of the proposed approach.
\end{itemize}

\section{Related Work}
In this section, we review the related works that are most relevant to us, including license plate recognition, natural scene text recognition and generative adversarial networks.

\subsection{License Plate Recognition} 
Existing license plate recognition algorithms can be divided into two categories: segmentation based methods~\cite{gou2015vehicle,guo2008license} and segmentation-free based methods~\cite{li2016reading}. The segmentation-based methods need to segment the license plate into individual characters, and then recognize them one by one. After the license plate segmentation is completed, template matching~\cite{rasheed2012automated} and learning based~\cite{wen2011algorithm} algorithms are usually used to classify characters. However, the segmentation methods lose the internal information of license plates, and the segmentation performance would seriously affect the recognition accuracy. Li and Shen~\cite{li2016reading} propose a cascaded framework based on CNN and LSTM for segmented free-based license plate recognition, which significantly improved the accuracy of standard license plate recognition. 
However, different from standard license plate, the enlarged license plate is with high diversity in both background and text, and existing methods can not handle the recognition of enlarged license plate well.

\subsection{Scene Text Recognition}

In recent years, many works have emerged to solve the task of irregular text recognition. Yang et al.~\cite{yang2017learning} and Li et al.~\cite{li2019show} use the two-dimensional attention mechanism for irregular text recognition, Liao et al.\cite{liao2019scene} use a two-dimensional perspective with a semantic segmentation network to identify irregular scene text. In addition, Luo et al.\cite{luo2019moran} propose a rectification network to convert irregular text images into regular text images that reduce background interference. Yang et al.\cite{yang2019symmetry} use character-level supervision in order to be able to train the model accurately. These methods improve the recognition accuracy of irregular text. 
However, huge data support is the key to training these network frameworks and building a large-scale training data is very time consuming and high labor cost. 

\subsection{Generative Adversarial Networks}

With the widespread application of the GAN network, Azadi et al.~\cite{azadi2018multi}, Cheng et al.~\cite{cheng2019structure}, and Yang et al.~\cite{yang2019controllable} have achieved amazing results on document images using adversarial learning methods. These methods focus on a single character and constrain the character to generate a single style. However, our goal is to generate enlarged license plates with various styles. This requires us to maintain character information while being able to generate a good background. The traditional binarization method works well on document images, but can not maintain the effect when the appearance of text in natural images changes greatly and there is noise. Therefore, it is still an open issue to coordinate the image generation of the content and the background.

Recently, several attempts in image translation have achieved a critical step. Isola et al.~\cite{isola2017image} achieve the generation of complex image pairs by using paired data sets and using pixel-level losses to generate complex image pairs. The CycleGAN model proposed by Zhu et al.~\cite{CycleGAN2017} and the cycle- consistency loss is used to solve the problem of unpaired data. The DRIT proposed by Lee et al.~\cite{DRIT_plus} uses different encoders to complete different tasks, and realizes the diverse generation of complex images by constraining the embedding space of different encoders. In order to make better use of the discriminator, Chen et al.~\cite{Chen_2020_CVPR} proposef the idea of reusing the discriminator encoder to generate more complex images with better quality. However, these methods~\cite{liu2017unsupervised, huang2018multimodal,yi2017dualgan,kim2017learning}, still have challenges in the generation of enlarged license plate.

These models can easily lose the detailed information of during the training process, which will have a negative impact on the recognizer. 
At the same time, some advanced image generation models~\cite{karras2019style,karras2018style,choi2018stargan,brock2018large,shao2021spatchgan,liu2022towards, zhou2022pro} can generate better synthetic images, however, these methods are unsupervised, and we can't use them to generate enlarged license plates of specified text.

In order to solve the problem of enlarged license plate insufficient data, we propose a novel task-level disentanglement generation framework based on the Disentangled Generation Network (DGNet),
which disentangles the generation into the text generation and background generation in an end-to-end manner to effectively ensure the diversity and integrity, for robust enlarged license plate recognition.

\section{DGNet: Disentangled Generation Network}
\begin{figure*}
\begin{center}
\includegraphics[width=\linewidth]{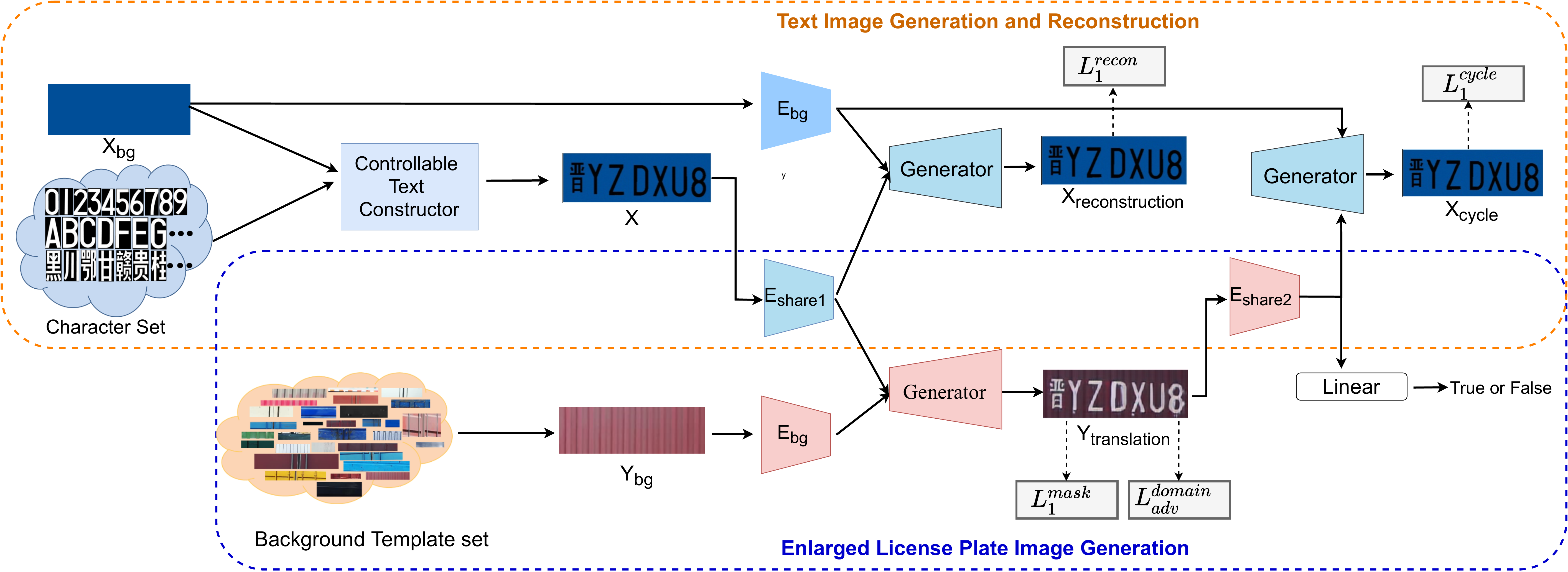}
\end{center}

\caption{
Pipeline of the proposed framework. $X$ is the text image, $X_{bg}$ is unified background image, and $Y_{bg}$ is the background template sampled from the background template set. $X_{reconstruction}$ is the reconstructed image by combination of text image and unified background image, $Y_translated$ is the synthetic enlarged license plate, and $X_{cycle}$ is the synthetic text image. $E_{bg}$ is the background encoder, and the discriminator consists of a shared encoder $E_{shared2}$ and a linear classifier.
}

\label{fig:overview}
\end{figure*}

In this section, we first introduce a novel task-level  disentanglement generation framework based on the  Disentangled Generation Network (DGNet), which disentangles the generation into the text generation and
background generation, for robust enlarged license plate recognition. Then we present a detailed description of mask constraint loss, which effectively ensures the integrity of generated enlarged license plates.

\subsection{Overview}
Inspired by NICEGAN~\cite{Chen_2020_CVPR} and DRIT~\cite{DRIT_plus}, we propose DGNet as shown in Fig.~\ref{fig:overview}. 
DGNet consists of four parts, including text image generation, background template set construction, enlarged license plate image generation and text image reconstruction.

Text image generation and background template set construction help us to generate highly diverse text images and background templates. Through this way, we can guarantee the diversity of generated enlarged license plates. 
Then, we use the text image and background template to generate the enlarged license plate by the module of enlarged license plate image generation, in which the mask constraint loss is introduced to ensure the integrity of the enlarged license plate. 
The problem of unpaired data is solved through the module of text image reconstruction.

In addition, inspired by NICEGAN~\cite{Chen_2020_CVPR}, our discriminator consists of a shared encoder $E_{share2}$ and a linear classifier. The shared encoder $E_{share2}$ shortens the domain translation path between low-dimensional hidden space vectors and promotes the domain translation between high-dimensional images.

\begin{figure}[t]
\begin{center}
\includegraphics[width=\linewidth]{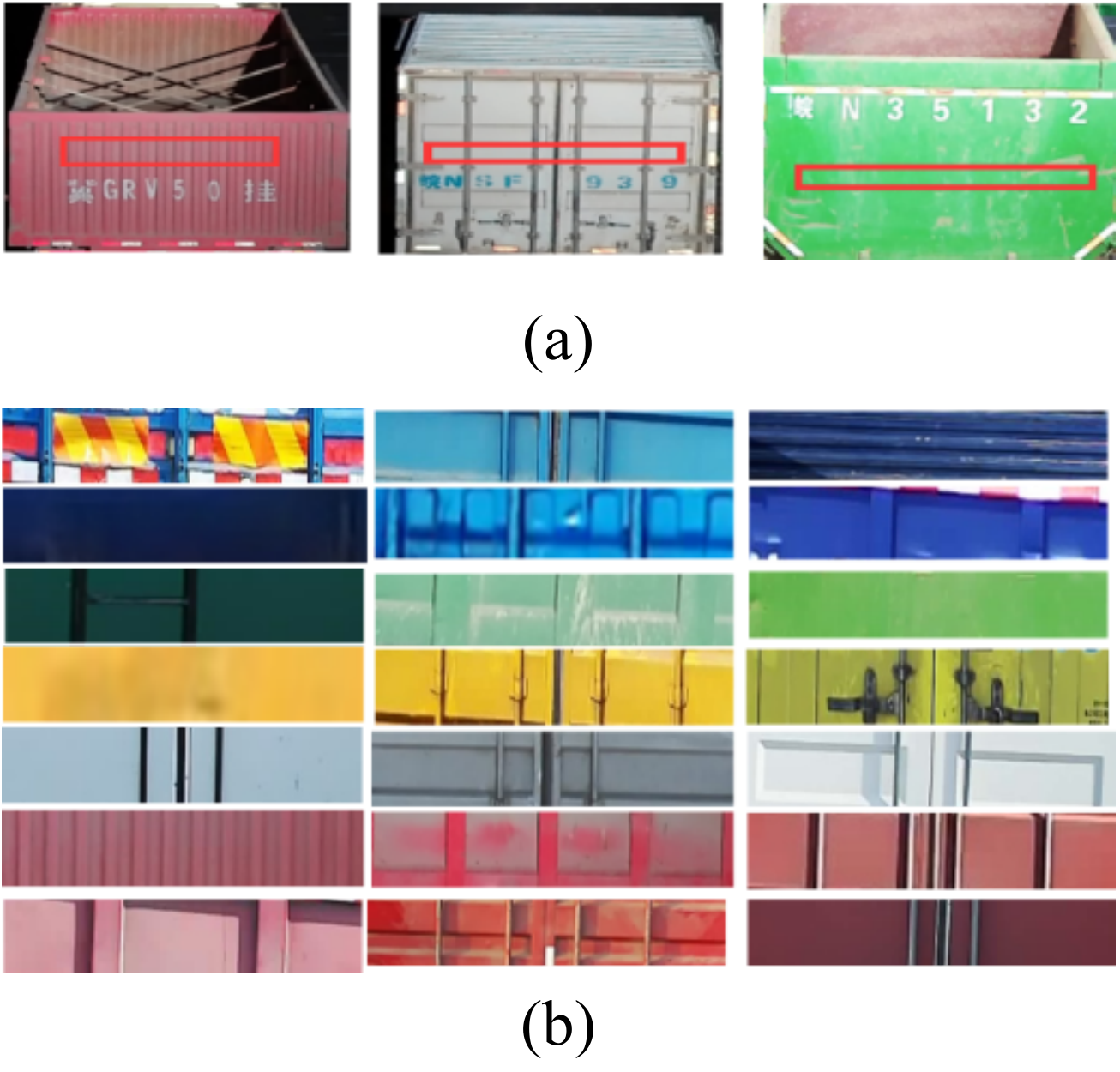}
\end{center}
\caption{
(a) Visualization of constructed background templates. Herein, we randomly select the background templates around the enlarged license plates. (b) Some samples from the background template set we built, which contains more than 1000 randomly cropped images from different regions and is highly diverse.
}
\label{fig:bg}
\end{figure}

\subsection{Controllable Text Image Generation}
As shown in Fig.~\ref{fig:challenge} (b), the characters of enlarged license plates lack uniform standards in font, position, and other attributes. 
To simulate real scenarios, we design a strategy of controllable text image generation.

In specific, we build a character set including all characters that existed in enlarged license plates, and combine some characters with a unified blue background image $X_{bg}$ by a controllable text constructor to form the text image $X$. 
In the controllable text constructor, we augment these character images by changing their sizes, shapes, and positions according to our requirements as follows.
First, the character images are randomly resized between $30 \times 60$ and $60 \times 120$. Second, to reshape these text images, we make the characters bigger by randomly expanding 0 to 2 pixels outward or make the characters smaller by randomly shrinking 0 to 2 pixels inward along the character edges. Finally, to simulate the character positions in real-world scenarios, we preset many sets of character positions, and the controllable text constructor will randomly choose from them. 
More importantly, the changes of these characters are completely retained in synthesized enlarged license plates, and the diversity is thus enhanced. 

\begin{figure*}
\begin{center}
\includegraphics[width=\linewidth]{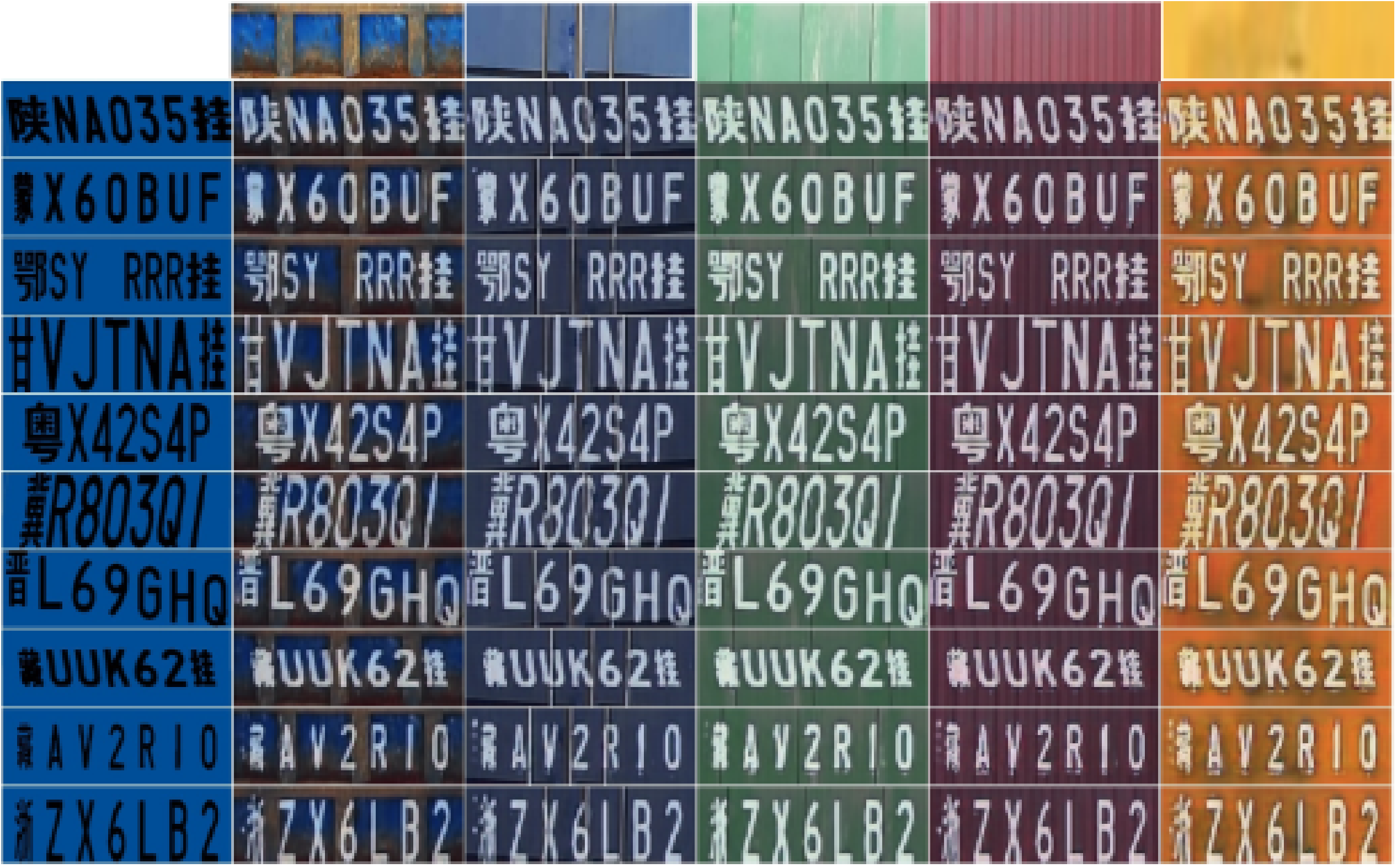}
\end{center}

\caption{
Visualization of synthetic enlarged license plates with different text images and background templates. The first row indicates different background templates and the first column denotes different text images. Other images are the synthetic enlarged license plates generated by our DGNet. 
}

\label{fig:display1}
\end{figure*}

As shown in Fig~\ref{fig:display1}, we can see that
the background style of the synthetic enlarged license plate can be controlled by different background templates, and the sizes and positions can be controlled by the augmented text images. In this way, we can generate enlarged license plates with controllable sizes, shapes, and positions.

\subsection{Multi-style Background Image Construction}
The complexity of background in enlarged license plates is an important factor that affects the performance of recognition methods. Existing works~\cite{DRIT_plus,huang2018multimodal} usually extract text and background information from a single image. However, enlarged license plates have high diversity and lots of noisy information, which makes it difficult to accurately separate the text and background from an image. 
Therefore, to better simulate the diverse background information, we propose to build a multi-style background template set. 
In specific, we extract the pure background template without character information around the enlarged license plates, as shown in Fig.~\ref{fig:bg} (a). The background template set contains more than 1000 randomly cropped images from different regions, and has various styles of background, which cover almost all styles in the real scenes, as shown in Fig.~\ref{fig:bg} (b). The background template can be randomly selected and combined with the text image to generate an enlarged license plate with the high-diversity background.

\subsection{Enlarged License Plate Image Generation}

Given the text image $X$ and the background template $Y_{bg}$, we first extract text and background features
by a two-stream encoder.
The shared encoder $E_{share1}$ is used to extract the text features, and the background encoder $E_{bg}$ is used to extract the background features of the input enlarged license plate. Then, we send the combined text and background features to the generator to generate the  high-quality synthesized enlarged license plate $Y_{translation}$. 

Due to the imbalance of training data, the generator often generates frequently seen characters in training data. Therefore, we introduce the mask constraint loss to ensure the integrity of generated enlarged license plate. In specific, we use a mask image to supervise the generation by introducing a mask constraint loss, which is calculated by the average absolute error between the mask image and the output image. Herein, the mask image is obtained by  subtracting the blue background image from the text image.
In addition, the features extracted by the shared encoder $E_{share2}$ are used as the inputs of text image reconstruction and the linear classifier.

\subsection{Text Image Reconstruction}

To solve the problem of unpaired training data, we introduce the text image reconstruction module.
On one hand, the cycle consistency loss~\cite{CycleGAN2017} is used to
force the images between two domains to complete generation with each other,
which effectively solve the unpaired training data problem. 
On the other hand, we combine the text features extracted by the shared encoder $E_{share1}$ with the background features extracted by the background encoder $E_{bg}$ as the input of the text image generator to get the reconstructed text image $X_{reconstruction}$ and the generated text image $X_{cycle}$. The text features of the $X_{reconstruction}$ come directly from the shared encoder directly connected to the text image, and the $X_{cycle}$ comes indirectly from this shared encoder. 

\subsection{Loss Function}
The overall loss consists of four parts, including generation adversarial loss, Cycle consistency loss, Reconstruction loss and mask constraint loss.

{\flushleft \bf Generation adversarial loss}.
First, we make use of the least-square adversarial loss by Mao et al.~\cite{mao2017least} for its more stable training and high-quality generation. The generation adversarial loss is the key to improve the performance of the generator, which can be represented as follows:
\begin{equation}
\begin{split}
    L^{domain}_{adv} = \log D_Y(Y) +\log(1-D_Y(G_{X\xrightarrow{}Y}(X,Y_{bg}))
\end{split}
\end{equation}
where $X$ is the text image, $Y$ is the enlarged license plate, $Y_{bg}$ is the background template,  $G_{X\xrightarrow{}Y}$ is the generator of text image to enlarged license plate, and $D$ is Discriminator.

{\flushleft \bf Cycle consistency loss}.
Inspired by CycleGAN~\cite{CycleGAN2017}, we use the cycle consistency loss to solve unpaired data problem. Cycle-consistency loss could to force the generators to be each others inverse and it is expressed as follows:
\begin{equation}
Y_{translation} = G_{X\xrightarrow{}Y}(E_{share1}(X),E_{bg}(Y_{bg})),
\end{equation}

\begin{equation}
L^{cycle}_{1} =|X - G_{Y\xrightarrow{}X}(E_{share2}(Y_{translation}),E_{bg}(X_{bg})) |_1,
\end{equation}
where $|\cdot|$ denotes the $l_1$ norm, $E_{share2}$ is the shared encoder, $G_{Y\xrightarrow{}X}$ is the generator of enlarged license plate to text image, $E_{bg}$ is the background encoder, $X_{bg}$ is the unified blue background image, and $Y_{translation}$ is the synthesized enlarged license plate.

{\flushleft \bf Reconstruction loss}.
In order to maintain the consistency of the generated background, we also introduce the reconstruction loss. Our reconstruction is based on the shared-latent space assumption~\cite{Chen_2020_CVPR}.
Reconstruction loss is to regularize the translation to be near an identity mapping when 
real samples’ hidden vectors of the source domain are provided as the input to the generator of the source domain. It is expressed as follows:
\begin{equation}
L^{recon}_{1} =|X-G_{Y\xrightarrow{}X}(E_{share1}(X),E_{bg}(X_{bg}))|_1,
\end{equation}
where  $E_{share1}$ is the shared encoder.

{\flushleft \bf Mask constraint loss}.
In order to effectively ensure the integrity of generated enlarged license plate, we design the mask constraint loss. It is expressed as follows:
\begin{equation}
\begin{split}
    L^{mask}_{1} = |I_{mask} - Y_{translation}|_1
\end{split}
\end{equation}
where $I_{mask}$ is the mask image used to supervise the generation.

{\flushleft \bf Full objective}.
The objective function is given by
\begin{equation}
    \begin{split}
        L =  \lambda_1 \cdot L^{domain}_{adv} + \lambda_2 \cdot L^{recon}_1 
        + \lambda_3 \cdot L^{cycle}_1 + \lambda_4 \cdot L^{mask}_1
    \end{split} 
\end{equation}
where $\lambda_1$, $\lambda_2$, $\lambda_3$ and $\lambda_4$ depicts a hyper-parameter used to balance
the trade-off between different supervisions, which are empirically set to $1.0$, $10.0$, $10.0$, $15.0$ respectively.

\section{ELPR Benchmark Dataset}
\begin{figure}[t]
\begin{center}
\includegraphics[width=\linewidth]{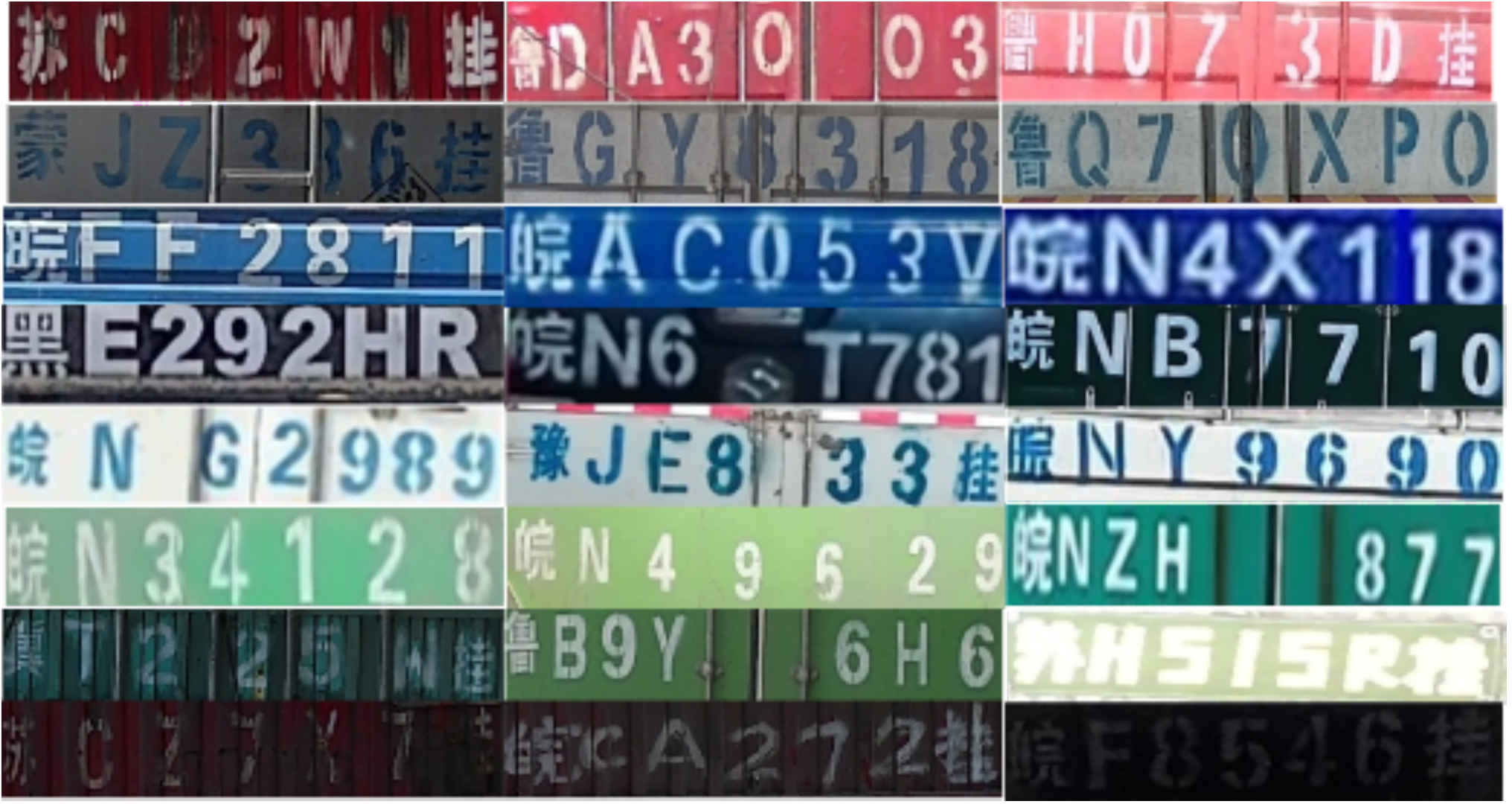}
\end{center}
\caption{
Examples from ELPR Dataset. The license plates come
from 18 different provinces, and all of them are captured
from real traffic monitoring scenes.
}
\label{fig:datasetsamples}
\end{figure}

\begin{figure}[t]
\begin{center}
\includegraphics[width=\linewidth]{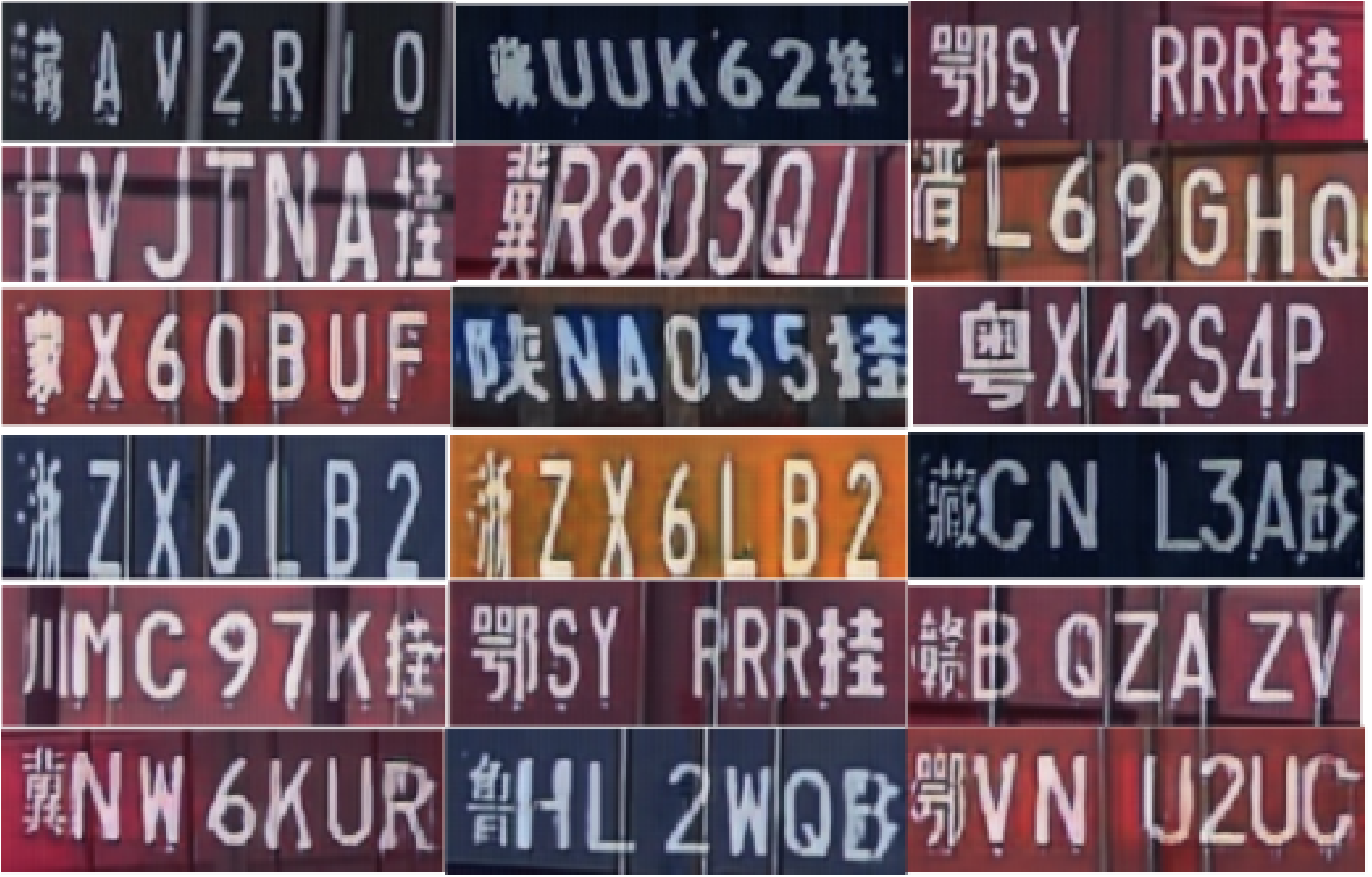}
\end{center}
\caption{
Examples of synthetic images. The proposed image generation approach can effectively maintain text information and generate clear background information by the mask constraint loss.
}
\label{fig:syc_datasetsamples}
\end{figure}

A large-scale dataset is crucial in enlarged license plate recognition because it can be used not only to train deep recognition models, but also to evaluate different recognition algorithms. Therefore, we provide a unified dataset for enlarged license plate recognition, called ELPR. We will introduce ELPR in detail.

\subsection{Data Creation}

The current recognition field of enlarged license plate recognition lacks a unified dataset, and we thus create a large-scale ELPR dataset. Our goal is to provide a unified and highly diverse dataset to cover real-world scenes and challenges. Therefore, we use surveillance cameras in real traffic scenes to capture enlarged license plates of large vehicles. Because the data comes from real scenes, our ELPR contains most real challenges such as occlusion, abnormal illumination, and blur. Fig.~\ref{fig:datasetsamples} shows typical examples of the ELPR dataset, and we can see that many factors, such as complex background and various text styles, increase the difficulty of ELPR. By the way, we collect a total of 9342 images, including the license plates of 18 different provinces. 
\begin{table*}
\caption{
Descriptions of 10 different challenges in ELPR dataset.
}
\small
\centering
\begin{tabular}{ll}
\toprule
\bf{Challenge} & \bf{Definition}  \\
\midrule
\bf{Inclined Angle} & The images of enlarged license plates are with different inclination angles. \\
\bf{Abnormal Illumination} & The images are captured in high or low illumination conditions. \\
\bf{Different Spacing}  & In a enlarged license plate, the spaces of different characters is different. \\
\bf{Size Variation}     &  In a enlarged license plate, the sizes of different characters are different. \\
\bf{Blur}  & The images are blur caused by fast motion and inaccurate camera focus. \\
\bf{Abrasion}     & The images are abraded due to the long-term use of vehicles, and some information of characters is missing. \\
\bf{Background Clutter}  & The background contains many noises which disturbs the recognition of enlarged license plate. \\
\bf{Non-standard Character}& The characters are handwritten instead of standard printed. \\
\bf{Double-row Plate}   & The enlarged license plate is painted in two rows. \\
\bf{Occlusion}& Some characters are partially or completely occluded by other objects. \\
\bottomrule
\\ 
\end{tabular}

\label{tab:Challenge_description}
\end{table*}
\begin{table}
\caption{
Distribution of different challenges in the ELPR dataset
}
\centering
\begin{tabular}{ll}

\toprule
  Challenge & Number\\
\midrule
  Inclined Angle & 590 \\ 
  Abnormal Illumination & 807  \\
  Different Spacing& 1492 \\
  Size Variation & 258 \\
  Blur& 650 \\
  Abrasion& 1404\\ 
  Background Clutter& 2435\\ 
  Non-standard Character& 485\\ 
  Double-row Plate& 10 \\
  Occlusion& 589 \\
\bottomrule
\\ 
\end{tabular}

\label{tab:Challenge_number}
\end{table}

\subsection{Annotation}
In order to ensure high-quality data, we train two professional annotators to label enlarged license plates one by one. In addition, we also ask professional checkers to prevent errors and inaccurate annotations.
In addition to the special challenges of enlarging license plate including complex background and high diversity in text size and position, etc., some common challenges such as abnormal illumination, blur, and occlusion also seriously affect the recognition performance. In order to better evaluate the performance of different recognition algorithms, we annotate each image with several challenges from the total 10 challenges, including 
inclined angle,  abnormal illumination, different spacing, size variation, blur, abrasion, background clutter, 
non-standard character, double-row plate, and occlusion. The challenges are defined in Table~\ref{tab:Challenge_description}, and Table~\ref{tab:Challenge_number} shows the number distribution of challenges of the ELPR dataset.

\subsection{Data Split}

There is no other dataset for enlarged license plate recognition, and thus we divide it into a training set and testing set to facilitate the training and evaluation of recognition methods. In specific, inspired by the standard license plate dataset CCPD~\cite{xu2018towards}, the testing set is randomly sampled, accounting for about 20~\% of the whole dataset.

\section{Evaluation}

In this section, we will provide the details of experiments and report the experimental results on the benchmark dataset ELPR to validate the effectiveness of our DGNet against the state-of-the-art methods.

\subsection{Evaluation Metrics}
Like some GAN evaluation methods, we adopt the Kernel Inception Distance (KID)~\cite{binkowski2018demystifying} and the Frechet Inception Distance (FID)~\cite{heusel2017gans} to evaluate the quality of image generation. FID compares the statistics of generated data against real data, and fits a Gaussian distribution to the hidden activations of InceptionNet for each compared image set and then computes the Frchet distance between those Gaussians. Lower FID is better, corresponding to more realistic generated images. KID is a metric similar to FID but uses the squared maximum mean discrepancy between Inception representations with a polynomial kernel. Unlike FID, KID has a simple unbiased estimator, making it more reliable especially when there are much more inception feature channels than image numbers. Lower KID indicates high visual similarity between real and generated images. 

In addition, our goal is to improve the recognition accuracy of enlarged license plates. Therefore, we adopt two extra indicators of enlarged license plate recognition accuracy (RA) and character recognition accuracy (CRA) for qualitative evaluation. Enlarged license plate recognition accuracy can be defined as: 
\begin{equation}
    RA = \frac{Number \ of \ correctly \ recognized\ license\ plates}{Number\ of\ all\ license\ plates},
\end{equation}
while character recognition accuracy can be defined as:
\begin{equation}
    CRA = \frac{Number\ of\ correctly\ recognized\ characters}{Number\ of\ all\ characters}.
\end{equation}

\begin{table}
\caption{ Comparison of three representative recognition methods with the synthetic images generated by different methods. 
} 
\centering
\begin{tabular}{clcccc}
\toprule
\multirow{3}{*}{Baseline}& \multirow{3}{*}{GAN Method} &\multicolumn{2}{c}{Training Data}  & \multirow{3}{*}{RA} & \multirow{3}{*}{CRA} \\
\cmidrule{3-4}
& &Real & Synthetic \\
\midrule
\multirow{4}{*}{Litman et al.\cite{litman2020scatter}}
& Script   &7K & 20K  & 76.82 & 94.40 \\
& NICEGAN\cite{Chen_2020_CVPR}  &7K & 20K  & 76.65 & 93.90 \\
& CycleGAN\cite{CycleGAN2017} &7K & 20K  & 76.40 & 93.92 \\
& DGNet (Ours)     &7K & 20K  & \textbf{80.11} & \textbf{94.69} \\
 \midrule
\multirow{4}{*}{Yue et al.\cite{yue2020robustscanner}}
& Script   &7K & 20K  & 56.14 & 88.10 \\
& NICEGAN\cite{Chen_2020_CVPR}  &7K & 20K  & 57.79 & 88.47 \\
& CycleGAN\cite{CycleGAN2017} &7K & 20K  & 56.18 & 87.65 \\
& DGNet (Ours)     &7K & 20K  & \textbf{60.62} & \textbf{89.12} \\
\midrule 
\multirow{4}{*}{Baek et al.\cite{baek2019wrong}}
& Script   &7K & 20K  & 61.94 & 90.17 \\
& NICEGAN\cite{Chen_2020_CVPR}  &7K & 20K  & 55.57 & 88.18 \\
& CycleGAN\cite{CycleGAN2017} &7K & 20K  & 58.36 & 89.00 \\
& DGNet (Ours)     &7K & 20K  & \textbf{65.15} & \textbf{90.74} \\
\bottomrule
\\
\end{tabular}

\label{tab:compare_with_other_GAN_in_recognition}
\end{table}
\begin{table}
\normalsize
\caption{
FID and KID ×100 for different algorithms. Herein, the lower values are better.
}
\centering
\begin{tabular}{ccc}
\toprule
Method & FID \bf{↓} & KID $ \times 100$ \bf{↓}\\
\midrule
CycleGAN\cite{CycleGAN2017} & 258.55 & 30.59\\
NICEGAN\cite{Chen_2020_CVPR}  & 182.97 & 16.93\\
Script   & 162.52 & 15.20\\
DGNet (Ours)& \bf{64.72} & \bf{3.46}\\
\bottomrule
\\ 
\end{tabular}

\label{tab:compare_with_other_GAN_in_image_quality}
\end{table}

\subsection{Implementation Details}



Our network is implemented based on Pytorch and trained with a single Tesla P40 GPU.
We use the Adam optimizer to optimize the proposed network with the learning rate 0.0001 and ($\beta_1$, $\beta_2$) = (0.5, 0.999) on Tesla P40 trained over 100K iterations. For the inputs, we resize all
images to the size of $256 \times 256$. In addition, we compute the mean and standard deviation of all images in training set and to normalize the inputs. The batch size is set to 1, and the training speed is about 1.5 iterations per second. In testing, the generation of an enlarged license plate costs 2.0 ms on average.

\begin{figure*}[t]
\begin{center}
\includegraphics[width=\linewidth]{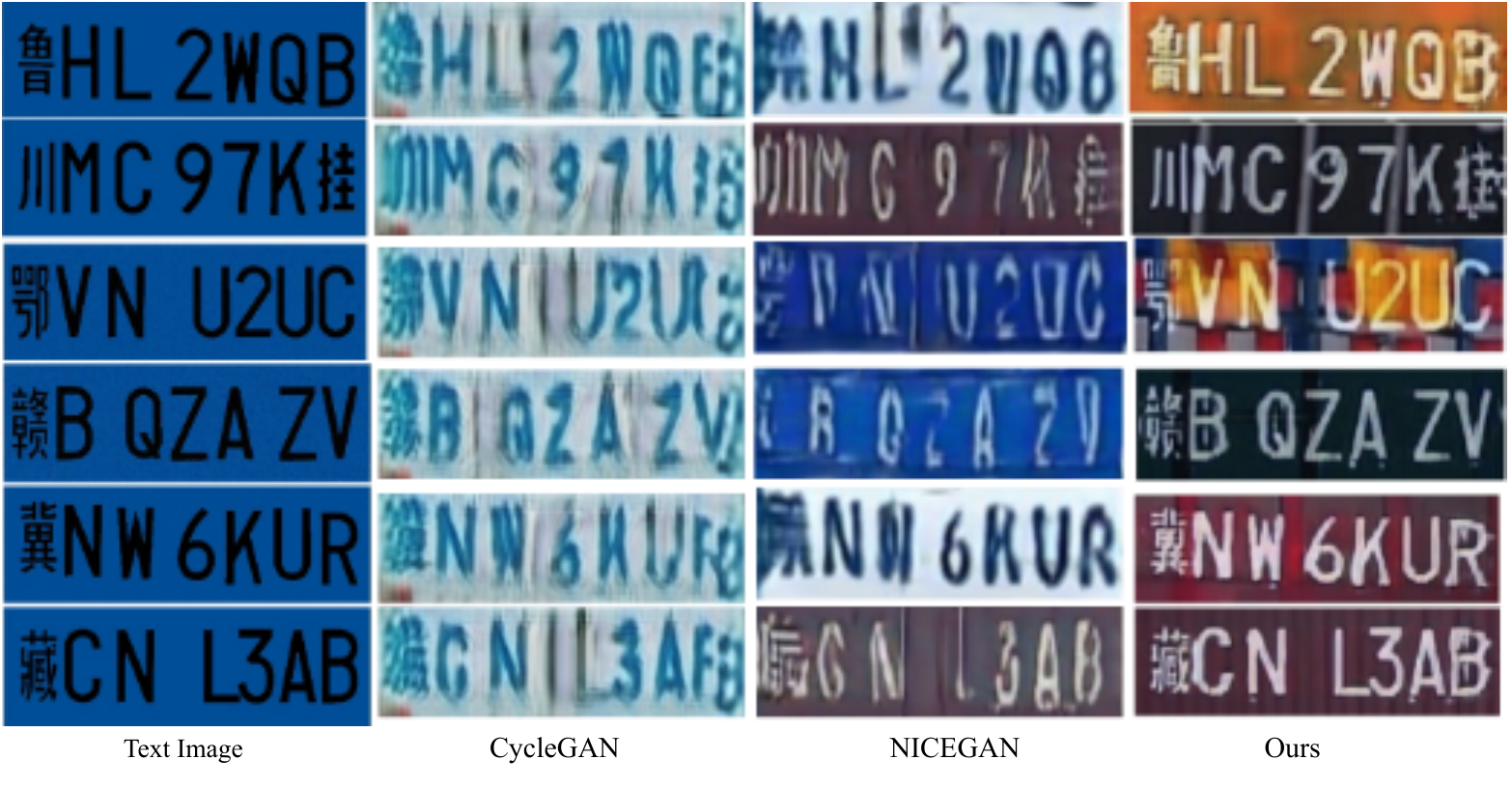}
\end{center}
\caption{
Comparison of the generated images by Cyclegan, NICEGAN, and our DGNet. 
}
\label{fig:compare}
\end{figure*}

\subsection{Comparison with State-of-the-art Methods}
In order to study the effectiveness of our synthesized enlarged license plates, we carry out experiments on several recent representative recognition algorithms, which are the models proposed by Litman et al.~\cite{litman2020scatter}, Yue et al~\cite{yue2020robustscanner} and Baek et al.~\cite{baek2019wrong} respectively. Litman et al. propose a modular four-stage scene text recognition framework, in which each component is interchangeable and allows different algorithms. On this basis, Yue et al. expand and propose that the stacked encoding and decoding modules achieve advanced performance. Baek et al. focuse on context-free text recognition with location attention, and show advanced performance in semantic-free text recognition.

\begin{table*}[]
\tiny
\caption{
Comparison results of RA/CRA scores(\%) of different recognition methods on different challenges in ELPR dataset, where R/S refers to real or synthetic data.
}
\begin{tabular}{llllllllllllllllllllllllll}
\toprule
Baseline      & Method        & \multicolumn{2}{l}{Training}  & \multicolumn{2}{l}{NOR}                  & \multicolumn{2}{l}{IA}                   & \multicolumn{2}{l}{AI}                   & \multicolumn{2}{l}{DS}                   & \multicolumn{2}{l}{SV}                   & \multicolumn{2}{l}{BLU}                  & \multicolumn{2}{l}{ABR}                  & \multicolumn{2}{l}{BC}                   & \multicolumn{2}{l}{NSC}                 & \multicolumn{2}{l}{DRP}     & \multicolumn{2}{l}{OCC}                  \\
              &               & \multicolumn{2}{l}{R/S} & \multicolumn{2}{l}{RA/CRA}               & \multicolumn{2}{l}{RA/CRA}               & \multicolumn{2}{l}{RA/CRA}               & \multicolumn{2}{l}{RA/CRA}               & \multicolumn{2}{l}{RA/CRA}               & \multicolumn{2}{l}{RA/CRA}               & \multicolumn{2}{l}{RA/CRA}               & \multicolumn{2}{l}{RA/CRA}               & \multicolumn{2}{l}{RA/CRA}              & \multicolumn{2}{l}{RA/CRA}  & \multicolumn{2}{l}{RA/CRA}               \\
\midrule
             
Litman et a~\cite{litman2020scatter}       & -             & \multicolumn{2}{l}{7K/0}      & \multicolumn{2}{l}{85.33/97.41}          & \multicolumn{2}{l}{82.35/96.64}          & \multicolumn{2}{l}{67.33/91.37}          & \multicolumn{2}{l}{78.03/94.8}           & \multicolumn{2}{l}{75.28/95.5}           & \multicolumn{2}{l}{66.19/90.88}          & \multicolumn{2}{l}{59.97/90.73}          & \multicolumn{2}{l}{70.71/93.52}          & \multicolumn{2}{l}{71.43/94.69}         & \multicolumn{2}{l}{\textbf{0/31.17}} & \multicolumn{2}{l}{59.11/90.41}          \\
              & CycleGAN      & \multicolumn{2}{l}{7K/20K}    & \multicolumn{2}{l}{88.8/97.99}           & \multicolumn{2}{l}{85.29/97.05}          & \multicolumn{2}{l}{71.29/91.56}          & \multicolumn{2}{l}{88.29/96.68}          & \multicolumn{2}{l}{82.02/96.79}          & \multicolumn{2}{l}{66.19/90.88}          & \multicolumn{2}{l}{60.28/90.05}          & \multicolumn{2}{l}{77.43/94.43}          & \multicolumn{2}{l}{69.52/94.29}         & \multicolumn{2}{l}{0/22.08} & \multicolumn{2}{l}{68.44/92.19}          \\
              & NICEGAN       & \multicolumn{2}{l}{7K/20K}    & \multicolumn{2}{l}{88.42/97.77}          & \multicolumn{2}{l}{\textbf{89.41/97.73}}          & \multicolumn{2}{l}{69.31/91.75}          & \multicolumn{2}{l}{84.52/95.40}           & \multicolumn{2}{l}{70.79/95.02}          & \multicolumn{2}{l}{69.52/90.82}          & \multicolumn{2}{l}{62.42/90.75}          & \multicolumn{2}{l}{76.49/93.74}          & \multicolumn{2}{l}{73.77/95.10}          & \multicolumn{2}{l}{0/25.97} & \multicolumn{2}{l}{69.78/92.51}          \\
              & Script        & \multicolumn{2}{l}{7K/20K}    & \multicolumn{2}{l}{89.38/98.12}          & \multicolumn{2}{l}{83.53/96.97}          & \multicolumn{2}{l}{67.00/91.47}             & \multicolumn{2}{l}{85.36/96.32}          & \multicolumn{2}{l}{71.91/95.50}           & \multicolumn{2}{l}{\textbf{70.48/91.77}} & \multicolumn{2}{l}{63.34/91.50}           & \multicolumn{2}{l}{78.17/\textbf{95.04}}          & \multicolumn{2}{l}{76.19/95.78}         & \multicolumn{2}{l}{0/29.22} & \multicolumn{2}{l}{66.22/92.70}           \\
              & DGNet (Ours) & \multicolumn{2}{l}{7K/20K}    & \multicolumn{2}{l}{\textbf{92.86/98.43}} & \multicolumn{2}{l}{88.24/ \textbf{97.73}} & \multicolumn{2}{l}{\textbf{71.95/92.27}} & \multicolumn{2}{l}{\textbf{89.33/96.56}} & \multicolumn{2}{l}{\textbf{83.15/97.43}} & \multicolumn{2}{l}{70.00/91.29}             & \multicolumn{2}{l}{\textbf{65.64/91.56}} & \multicolumn{2}{l}{\textbf{80.97}/94.94} & \multicolumn{2}{l}{\textbf{78.10/95.92}} & \multicolumn{2}{l}{0/24.68} & \multicolumn{2}{l}{\textbf{70.67/93.46}} \\
\midrule
Yue et al.~\cite{yue2020robustscanner} & -             & \multicolumn{2}{l}{7K/0}      & \multicolumn{2}{l}{17.95/71.4}           & \multicolumn{2}{l}{15.29/66.39}          & \multicolumn{2}{l}{10.23/62.47}          & \multicolumn{2}{l}{9.00/63.09}              & \multicolumn{2}{l}{12.36/65.33}          & \multicolumn{2}{l}{14.76/65.1}           & \multicolumn{2}{l}{11.96/64.48}          & \multicolumn{2}{l}{8.77/62.39}           & \multicolumn{2}{l}{14.29/67.07}         & \multicolumn{2}{l}{\textbf{0/24.03} } & \multicolumn{2}{l}{10.22/62.6}           \\
              & CycleGAN      & \multicolumn{2}{l}{7K/20K}    & \multicolumn{2}{l}{71.82/94.21}          & \multicolumn{2}{l}{63.53/91.01}          & \multicolumn{2}{l}{44.22/82.74}          & \multicolumn{2}{l}{59.41/88.76}          & \multicolumn{2}{l}{55.06/89.73}          & \multicolumn{2}{l}{50.48/83.67}          & \multicolumn{2}{l}{44.79/82.87}          & \multicolumn{2}{l}{53.92/86.89}          & \multicolumn{2}{l}{43.81/86.39}         & \multicolumn{2}{l}{0/18.83} & \multicolumn{2}{l}{43.11/85.59}          \\
              & NICEGAN       & \multicolumn{2}{l}{7K/20K}    & \multicolumn{2}{l}{73.17/95.01}          & \multicolumn{2}{l}{63.94/91.85}          & \multicolumn{2}{l}{46.21/83.22}          & \multicolumn{2}{l}{62.55/\textbf{90.71} }          & \multicolumn{2}{l}{57.30/91.01}           & \multicolumn{2}{l}{48.57/84.97}          & \multicolumn{2}{l}{46.47/84.98}          & \multicolumn{2}{l}{55.03/87.50}           & \multicolumn{2}{l}{53.55/87.48}         & \multicolumn{2}{l}{0/18.18} & \multicolumn{2}{l}{44.89/85.90}           \\
              & Script        & \multicolumn{2}{l}{7K/20K}    & \multicolumn{2}{l}{72.59/94.68}          & \multicolumn{2}{l}{56.47/90.59}          & \multicolumn{2}{l}{44.22/82.46}          & \multicolumn{2}{l}{60.46/89.12}          & \multicolumn{2}{l}{56.18/91.49}          & \multicolumn{2}{l}{\textbf{54.29/85.51}} & \multicolumn{2}{l}{42.64/83.87}          & \multicolumn{2}{l}{52.61/86.83}          & \multicolumn{2}{l}{\textbf{58.10/88.57}} & \multicolumn{2}{l}{0/22.73} & \multicolumn{2}{l}{41.78/84.63}          \\
              & DGNet (Ours) & \multicolumn{2}{l}{7K/20K}    & \multicolumn{2}{l}{\textbf{76.64/95.26}} & \multicolumn{2}{l}{\textbf{65.88/92.86}} & \multicolumn{2}{l}{\textbf{51.16/84.77}} & \multicolumn{2}{l}{\textbf{65.06}/90.68} & \multicolumn{2}{l}{\textbf{65.17/92.46}} & \multicolumn{2}{l}{50.00/85.17}             & \multicolumn{2}{l}{\textbf{46.78/84.40}}  & \multicolumn{2}{l}{\textbf{59.52/88.83}} & \multicolumn{2}{l}{49.52/87.48}         & \multicolumn{2}{l}{0/17.53} & \multicolumn{2}{l}{\textbf{46.67/86.98}} \\
\midrule
Baek et al.~\cite{baek2019wrong}      & -             & \multicolumn{2}{l}{7K/0}      & \multicolumn{2}{l}{66.41/92.2}           & \multicolumn{2}{l}{56.47/89.16}          & \multicolumn{2}{l}{42.9/82.84}           & \multicolumn{2}{l}{52.09/86.28}          & \multicolumn{2}{l}{50.56/88.93}          & \multicolumn{2}{l}{53.81/85.85}          & \multicolumn{2}{l}{42.95/83.11}          & \multicolumn{2}{l}{50.37/85.79}          & \multicolumn{2}{l}{48.57/88.57}         & \multicolumn{2}{l}{\textbf{0/22.73} } & \multicolumn{2}{l}{37.78/82.67}          \\
              & CycleGAN      & \multicolumn{2}{l}{7K/20K}    & \multicolumn{2}{l}{75.68/95.12}          & \multicolumn{2}{l}{65.88/92.52}          & \multicolumn{2}{l}{46.21/84.87}          & \multicolumn{2}{l}{63.39/90.38}          & \multicolumn{2}{l}{57.3/91.65}           & \multicolumn{2}{l}{56.67/86.6}           & \multicolumn{2}{l}{43.25/83.7}           & \multicolumn{2}{l}{55.22/88.3}           & \multicolumn{2}{l}{49.52/89.52}         & \multicolumn{2}{l}{0/25.32} & \multicolumn{2}{l}{44/85.84}             \\
              & NICEGAN       & \multicolumn{2}{l}{7K/20K}    & \multicolumn{2}{l}{72.01/94,43}          & \multicolumn{2}{l}{66.47/92.94}          & \multicolumn{2}{l}{44.88/83.4}           & \multicolumn{2}{l}{59.41/89.15}          & \multicolumn{2}{l}{49.44/90.21}          & \multicolumn{2}{l}{50.95/85.17}          & \multicolumn{2}{l}{42.49/83.39}          & \multicolumn{2}{l}{54.29/87.87}          & \multicolumn{2}{l}{42.86/88.03}         & \multicolumn{2}{l}{0/27.92} & \multicolumn{2}{l}{39.11/82.6}           \\
              & Script        & \multicolumn{2}{l}{7K/20K}    & \multicolumn{2}{l}{77.22/95.64}          & \multicolumn{2}{l}{68.24/92.61}          & \multicolumn{2}{l}{\textbf{51.49} /85.81}          & \multicolumn{2}{l}{69.87/91.93}          & \multicolumn{2}{l}{\textbf{66.29} /92.3}           & \multicolumn{2}{l}{\textbf{59.05/88.91}} & \multicolumn{2}{l}{46.01/\textbf{85.45} }          & \multicolumn{2}{l}{60.08/89.69}          & \multicolumn{2}{l}{50.48/88.57}         & \multicolumn{2}{l}{0/28.57} & \multicolumn{2}{l}{\textbf{48.89} /87.49}          \\
              & DGNet (Ours) & \multicolumn{2}{l}{7K/20K}    & \multicolumn{2}{l}{\textbf{79.92/96.25}} & \multicolumn{2}{l}{\textbf{74.71/94.45}} & \multicolumn{2}{l}{\textbf{51.49/85.90}}  & \multicolumn{2}{l}{\textbf{74.48/92.62}} & \multicolumn{2}{l}{\textbf{66.29/93.58}} & \multicolumn{2}{l}{\textbf{59.05} /88.50}           & \multicolumn{2}{l}{\textbf{48.62} /85.43}          & \multicolumn{2}{l}{\textbf{65.30/91.15}}  & \multicolumn{2}{l}{\textbf{60.95/90.20}} & \multicolumn{2}{l}{0/25.33} & \multicolumn{2}{l}{45.78/\textbf{87.62} }     \\
\bottomrule
\label{tab:Attribute-based}
\end{tabular}
\end{table*}

\textbf{\flushleft Overall performance}.
We use different methods to generate synthetic data for experimental verification. 
First, to ensure the integrity and diversity of the generated enlarged license plate, we use image processing technology based on python to extract the character mask from the text image and add it to the background template to get the synthetic image for comparison. We name this scheme as Script method in our paper.
The second one is to synthesize enlarged license plates using existing GAN methods, including CycleGAN~\cite{CycleGAN2017} and NICEGAN~\cite{Chen_2020_CVPR}. 
The final one is to use our proposed model to generate enlarged license plates. We use three representative methods to train with the above synthetic data respectively and evaluate them on testing set. 

As shown in Table~\ref{tab:compare_with_other_GAN_in_recognition}, the results show that our method significantly improves the recognition accuracy of these recognition methods. In particular, our method is 3.46\%, 3.71\% and 3.29\% higher than NICEGAN~\cite{Chen_2020_CVPR}, CycleGAN~\cite{CycleGAN2017} and Script methods respectively. 
In addition, in the recognition framework proposed by Litman et al.~\cite{litman2020scatter} and Baek et al.~\cite{baek2019wrong}, Script method is 0.42~\%, 0.17~\% and 3.58~\%, 6.37~\% higer than CycleGAN~\cite{CycleGAN2017}  and NICEGAN~\cite{Chen_2020_CVPR} in RA, which can be explained by this method can generate higher diverse license plates than CycleGAN~\cite{CycleGAN2017} and NICEGAN~\cite{CycleGAN2017}. However, our DGNet can generate enlarged license plates with high diversity and small domain gap, therefore our method outperforms Script method in a large margin. 


Note that our synthesized data are particularly obvious for the recognition framework proposed by Yue et al.~\cite{yue2020robustscanner}, the overall performance in RA is 2.83\%, 4.44\%, 4.48\%  higher than NICEGAN~\cite{Chen_2020_CVPR}, CycleGAN~\cite{CycleGAN2017} and Script methods respectively.
This recognition method uses the positional attention enhancement to extract character features in images.
Training data is not enough to cover the position diversity of real scenes, which affects the performance of enlarged license plate recognition. 
while our synthesized data shows high diversity in position, size and other attributes and effectively makes up for the shortage of training data, which can improve the robustness of this recognition framework.

The evaluation of synthetic image quality is shown in  Table~\ref{tab:compare_with_other_GAN_in_image_quality}. The FID scores and KID scores of our proposed method are obviously smaller than those of CycleGAN~\cite{CycleGAN2017}, NICEGAN~\cite{Chen_2020_CVPR} and Script, which means that the enlarged license plates generated by our DGNet is more similar to real data.

The visualization of enlarged license plates generated by different methods is shown in Fig.~\ref{fig:compare},
It can be found that although the synthesized data generated by the Script have reliable character information, there is obvious domain difference with enlarged license plates.  CycleGAN~\cite{CycleGAN2017} and NICEGAN~\cite{Chen_2020_CVPR} lose the character structure information in the training process, which seriously affects the recognition performance. Note that the synthesized data generated by any method can improve the performance of the recognizer. It is because that even though their character structures have been destroyed and can not work at the pixel level, they can still effectively train recognition models with the feature-level information.

\begin{table}
\caption{
Comparison results of different methods using different number of real data, where the number of synthetic images is fixed.
}
\centering
\normalsize
\begin{tabular}{ccccc}
\toprule
\multirow{3}{*}{Method} &\multicolumn{2}{c}{Training Data}  & \multirow{3}{*}{RA} & \multirow{3}{*}{CRA} \\
\cmidrule{2-3}
&Real & Synthetic \\
\midrule
\multirow{4}{*}{Litman et al.\cite{litman2020scatter}}
&0 & 25K   & 6.86  & 55.21 \\
&2K & 20K  & 74.39 & 93.06 \\
&5K & 20K  & 78.13 & 94.45 \\
&7K & 20K  & \textbf{80.11}  & \textbf{94.69} \\
\midrule
\multirow{4}{*}{Yue et al.\cite{yue2020robustscanner}}
&0 & 25K   & 4.40  & 45.55 \\
&2K & 20K  & 50.10 & 85.10 \\
&5K & 20K  & 57.38 & 88.27 \\
&7K & 20K  & \textbf{60.62} & \textbf{89.12} \\
\midrule
\multirow{4}{*}{Baek et al.\cite{baek2019wrong}}
&0 & 25K   & 9.04 & 60.11 \\
&2K & 20K  & 59.68 & 88.72 \\
&5K & 20K  & 63.46 & 90.30 \\
&7K & 20K  & \textbf{65.15} & \textbf{90.74} \\
\bottomrule
\\
\end{tabular}

\label{tab:Real_data_of_different_proportion}
\end{table}

\begin{table}
\centering
\caption{
Comparison results of different setting in our method in different recognition methods, including without synthetic data and mask constraint loss respectively.
}
\scriptsize
\begin{tabular}{clcccc}
\toprule
\multirow{3}{*}{Baseline}& \multirow{3}{*}{Method}
&\multicolumn{2}{c}{Training Data}  & \multirow{3}{*}{RA} & \multirow{3}{*}{CRA} \\
\cmidrule{3-4}
& &Real & Synthetic \\
\midrule 
\multirow{3}{*}{Litman et al.\cite{litman2020scatter}}
& DGNet (Ours) & 7K & 0   & 72.13 & 93.42 \\
& DGNet w/o $ L_{mask}$ & 7K & 20K   & 74.56 & 94.02 \\
& DGNet (Ours)& 7K & 20K  & \textbf{80.11} & \textbf{94.69} \\
\midrule 
\multirow{3}{*}{Yue et al.\cite{yue2020robustscanner}}
& DGNet (Ours) & 7K & 0   & 12.82 & 65.82 \\
& DGNet w/o $ L_{mask}$ & 7K & 20K   & 51.38 & 86.03 \\
& DGNet (Ours)&7K & 20K  & \textbf{60.62} & \textbf{89.12} \\
\midrule 
\multirow{3}{*}{Baek et al.\cite{baek2019wrong}}
& DGNet (Ours) &7K & 0   & 52.53 & 86.68 \\
& DGNet w/o $ L_{mask}$ & 7K & 20K   & 48.66 & 86.00 \\
& DGNet (Ours)&7K & 20K  & \textbf{65.15} & \textbf{90.74} \\
\bottomrule \\
\end{tabular}

\label{tab:ablation}
\end{table}

\textbf{\flushleft Attribute-based performance}.
To analyze the performance under different challenges faced by existing recognition methods, we evaluate our algorithm against three algorithms on 11 challenge attributes including inclined angle (IA),  abnormal illumination (AI), different spacing (DS), size variation (SV), blur (BLU), abrasion (ABR), background clutter (BC), 
non-standard character (NSC), double-row plate (DRP) and occlusion (OCC), as shown in Table \ref{tab:Attribute-based}.
It can be seen from the results that our DGNet achieves better results in most attributes compared with all other image generation algorithms. Note that RA is more important evaluation metric. Under the attributes of IA, AI, DS, SV, ABR, and BC, our method is better than other methods in RA score, which further proves the diversity of our synthetic enlarged license plates and simulates the real data better.

There are two key points here need to be explained. First, under the attribute of DRP, the enlarged license plates are not recognized well, because these recognition methods are based on single-line text recognition. Second, although our method does not achieve the best results in all attributes, it performs best in most attributes, which further shows that our DGNet can generate highly diverse license plates. Other works~\cite{Chen_2020_CVPR, CycleGAN2017} tend to generate license plates with specific attributes due to mode collapse.

\begin{figure}[t]
\begin{center}
\includegraphics[width=\linewidth]{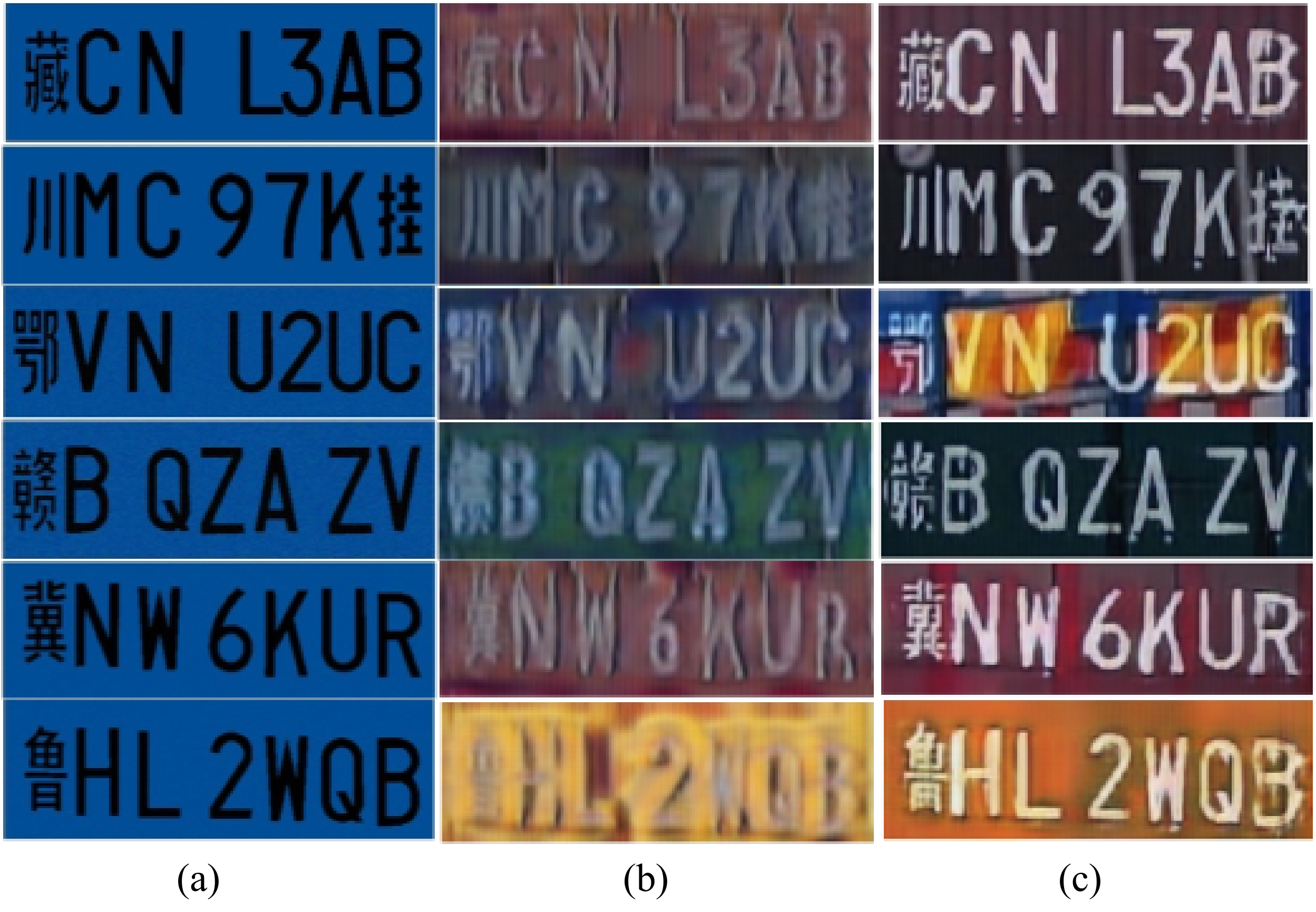}
\end{center}
\caption{
Comparison of the generated images using mask constraint loss. (a) Input text images. (b) The generated images without the mask constraint loss. (c) The generated images with the mask constraint loss. 
}
\label{fig:mask}
\end{figure}

\subsection{Ablation Study}
In order to further study the role of synthetic data, we set up an ablation experiment to  gradually increase the proportion of training data, including $0, 2k, 5K$ and $7K$ respectively, and evaluate them on testing data. As shown in Table~\ref{tab:Real_data_of_different_proportion}, when only synthetic data are used, the model can not work well. A reasonable explanation is that there is a obvious domain gap between synthetic data and real data. By increasing the proportion of real data step by step, the mixing of data will reduce the domain gap, so the recognition accuracy of the models will be continuously improved. Through these experiments, it can be found that our synthetic data can effectively improve the recognition performance of enlarged license plates.

To verify the effectiveness of the mask constraint loss, we set up an ablation experiment to compare the generated images without and with the mask constraint loss as shown in Fig.~\ref{fig:mask}. It can be found that the generated image quality is higher after adding the mask constraint loss, which proves that the mask constraint loss contributes greatly to the generated character quality and background. The mask constraint loss supervises the image at the pixel level, thus effectively maintains the structural information of text characters. As shown in Table~\ref{tab:ablation}, the mask constraint loss helps our DGNet improve 5.55\%, 9.34\%, and 16.49\% in RA scores respectively.

\section{Conclusion}
In this paper, we construct a unified enlarged license plate recognition dataset, which contains most of challenges in real scenes.
We also propose a task-level disentanglement generation
framework based on the disentangled generation network to effectively ensure the diversity and integrity of enlarged license plates, and thus greatly improving the recognition accuracy.
Extensive experiments on the dataset demonstrate the effectiveness of our method in different recognition frameworks. By releasing this dataset, we believe it will help the research and development of enlarged license plate recognition. 

Although we have explored the way of synthesizing enlarged license plates, there are still many potential problems to be solved. For example, the domain difference between synthesized enlarged license plates and real ones is large. In addition, how to use synthetic data to train the recognizer more efficiently and how to design an effective recognition model to address the unique challenges of enlarging license plate recognition are two unsolved issues. In the future, we will study more effective recognition algorithms and image generation algorithms to further improve the performance of enlarged license plate recognition, and expand the dataset to cover more  realistic scenes.

{
\bibliographystyle{IEEEtran}
\bibliography{reference}{}
}


 





\end{document}